\documentclass[3p,times]{elsarticle}

\usepackage{ecrc}
\usepackage{times}
\usepackage[backref]{hyperref}
\hypersetup{hypertex=true,
            colorlinks=true,
            linkcolor=blue,
            anchorcolor=blue,
            citecolor=blue}
\usepackage{graphicx}
\usepackage{algorithm}
\usepackage{algorithmic}

\usepackage{amsmath}
\usepackage{amssymb}
\usepackage{newfloat}
\usepackage{listings}
\usepackage{booktabs}
\usepackage{multicol}
\usepackage{multirow}
\usepackage{threeparttable}
\usepackage{array}
\usepackage{bm}
\usepackage{float}

\volume{00}

\firstpage{1}

\journalname{Neurocomputing}

\runauth{T. Yu et al.}

\jid{Neurocomputing}

\jnltitlelogo{Neuro- computing}

\usepackage{amssymb}
\usepackage[figuresright]{rotating}

\begin{document}
\begin{frontmatter}



\dochead{}
%% e.g. \dochead{17th International Conference on Dynamical Processes in Excited States of Solids}

\title{Recursive Least Squares for Training and Pruning \\ Convolutional Neural Networks}


\author{Tianzong Yu}
%\ead{20081200210007@hainanu.edu.cn}
\author{Chunyuan Zhang\corref{cor1}}
\ead{zcy7566@126.com}
\author{Yuan Wang}
%\ead{20085400210072@hainanu.edu.cn}
\author{Meng Ma}
\author{Qi Song}
\address{School of Computer Science and Technology, Hainan University, Haikou, Hainan 570228, China}
\cortext[cor1]{Corresponding author}
\begin{abstract}
Convolutional neural networks (CNNs) have succeeded in many practical applications.
However, their high computation and storage requirements often make them difficult to deploy on resource-constrained devices.
In order to tackle this issue, many pruning algorithms have been proposed for CNNs, but most of them can't prune CNNs to a reasonable level. In this paper, we propose a novel algorithm for training and pruning CNNs based on the recursive least squares (RLS) optimization.
After training a CNN for some epochs, our algorithm combines inverse input autocorrelation matrices and weight matrices to evaluate and prune unimportant input channels or nodes layer by layer. Then, our algorithm will continue to train the pruned network, and won't do the next pruning until the pruned network recovers the full performance of the old network.
Besides for  CNNs, the proposed algorithm can be used for feedforward neural networks (FNNs).
Three experiments on MNIST, CIFAR-10 and SVHN datasets show that our algorithm can achieve the more reasonable pruning and have higher learning efficiency than other four popular pruning algorithms.
\end{abstract}

\end{frontmatter}

% \linenumbers

\section{Introduction}
Convolutional neural networks (CNNs) are perhaps the most widely used network among deep neural networks (DNNs), which can extract sample features from different levels through unique convolutional and pooling mechanisms \cite{6795724}, so that they are particularly suitable for processing computer vision applications. However, they are generally very complex and require high computational and storage costs, which restricts their widespread application to a certain extent  \cite{DBLP:conf/cvpr/HeZRS16,DBLP:conf/iccv/HeGDG17}. In recent years, mobile devices, such as smart phones, wearables and drones, have been increasingly used, thus there remains a growing demand for deploying CNNs on these devices, which have much lower computational and storage capacity than conventional computers.
Thus, how to compress CNNs has become  one of research focus in deep learning.

Recently, there have been many model compression algorithms proposed for CNNs. In general, they can be divided into the following five categories  \cite{8253600}:
1) network pruning,
2) parameter quantization,
3) low-rank factorization,
4) filter compacting,
5) knowledge distillation.
In detail, network pruning algorithms prune redundant and uninformative channels or zero out unimportant weights  \citep{DBLP:journals/jetc/AnwarHS17,DBLP:journals/corr/abs-2003-03033}.
Parameter quantization algorithms reduce bits of all parameters for reducing computational and storage costs  \cite{DBLP:journals/ijon/LotricB12,DBLP:journals/corr/abs-1712-05877}.
Low-rank factorization algorithms decompose three-dimensional filters to two-dimensional filters  \cite{9361699}.
Filter compacting algorithms use compact filters to replace loose and over-parameterized filters  \cite{NIPS2016_b73dfe25}.
Knowledge distillation algorithms distill knowledge from the original network to generate a
small network \cite{DBLP:journals/corr/HintonVD15,DBLP:conf/iclr/ZhangM21}.
Compared with other categories, network pruning has received much more research attention.
Thus, we will focus on it in this paper.

Network pruning can be further divided into structured pruning and unstructured pruning.
The former generally removes the output channels which have less impact on the output loss of CNNs \cite{DBLP:conf/aaai/Hu0ZAL21}.
The latter directly zeroes out the negative or small weights in CNNs \cite{DBLP:conf/ijcai/HeKDFY18}.
In recent years, network pruning algorithms have been widely used for CNNs, but they still have some drawbacks.
Firstly, for ease of implementation, they generally use Momentum \cite{DBLP:journals/nn/Qian99} rather than Adam \cite{DBLP:journals/corr/KingmaB14}, RMSprop \cite{rmsprop} or Adadelta \cite{DBLP:journals/corr/abs-1212-5701} for CNNs, which leads to low training efficiency.
Secondly, they usually use one-shot pruning and their pruning ratio entirely depends on the manual setting, which
cannot automatically adapt to the CNN size and may lead to pruning too much or too little.
Thirdly, unstructured pruning does not reduce computational and storage costs substantially.
Lastly, they often prune nodes by analyzing weights or the loss change.
Different from traditional shallow  pruning algorithms, they don't use input features for pruning.

In a response to the above problems, based on the recursive least squares (RLS) optimization for CNNs \cite{Zhang}, we propose a novel training and pruning method. After training some epochs, our algorithm combines inverse input autocorrelation matrices and weight matrices to evaluate the importance of input channels (or nodes), and prunes the least important parts in each layer.  Then, it continues to train the pruned CNN, and don't perform the next pruning until the performance is fully recovered.
Compared with existing pruning algorithms, our algorithm has the following advantages:
1) Besides weight matrices, our algorithm also uses the inverse input autocorrelation matrices to measure the importance of input features, so it can prune CNNs according to the task's difficulty level adaptively.
2) Different from traditional one-shot pruning algorithms, which prune a large number of channels (or nodes) only one time and may cause an unrecovered impact on the pruned CNN, our algorithm prunes a CNN many times in the training process, so its pruning is more reasonable.
3) Different from existing algorithms only prune the channels (or nodes) in hidden layers, our algorithm can also prune the original features of training samples.
4) Our algorithm has faster training efficiency than previous algorithms since it uses the RLS optimization. Whereas other algorithms generally use Momentum optimization, resulting in lower training efficiency.
We compare our algorithm with four other popular network pruning algorithms on three benchmark datasets, MNIST, CIFAR-10 and SVHN. Experimental results show that our algorithm can prune CNNs with little loss, and can be used to prune feedforward neural networks (FNNs) as well. In addition, our algorithm has higher training efficiency and better pruning performance.

The rest of this  paper is organized as follows: In Section 2, the background of the RLS optimization algorithm is presented. In Section 3, we describe our proposed algorithm in detail. In Section 4, we discuss recently pruning algorithms and compare with our algorithm. Experiment results are shown in Section 5. Finally, we conclude our work in Section 6.

\section{Background}
In this section, we introduce the background knowledge used in this paper. We first review the RLS derivation, and then review the learning mechanism of CNNs with RLS optimization proposed by \cite{Zhang}. Some notations used in this paper are also introduced.
\subsection{Recursive Least Squares}
RLS is the recursive version of the linear least squares algorithm. It has fast convergence and is more suitable for online learning.
Suppose all sample inputs from the start to the current time are $\textbf X_t=\{\textbf x_1,\cdots,\textbf x_t\}$, and the corresponding expected outputs are $\textbf Y_t^* = \{y_1^*,\cdots,y_t^*\}$. Then, the least squares loss function is defined as
\begin{align}
J(\textbf w) = \frac{1}{2}\sum_{i=1}^t\lambda^{t-i}{(y_i^*-\textbf w^\textrm{T}\textbf x_i)}^2
\end{align}
where $\textbf w$ is the weight vector and $\lambda\in (0,1]$ is the forgetting factor.
Let $\nabla_\textbf wJ(\textbf w)=\bm{0}$, we can get
\begin{align}
 \sum_{i=1}^t\lambda^{t-i} (y_i^*-{{\textbf{w}}^*}^\textrm{T}\textbf x_i) \textbf x_i  = \bm{0}
\end{align}
Thus, we can obtain the least squares solution
\begin{align}
\textbf w_t = \textbf w^* = \textbf A_t^{-1}\textbf b_t
\end{align}
where $\textbf A_t$ and $\textbf b_t$ are defined as follows
\begin{align}
\textbf A_t =\sum_{i=1}^t\lambda^{t-i}\textbf x_i\textbf x_i^\textrm{T}=\lambda \textbf A_{t-1}+\textbf x_t\textbf x_t^\textrm{T}
\end{align}
\begin{align}
\textbf b_t =\sum_{i=1}^t\lambda^{t-i}\textbf x_iy_i^* =\lambda \textbf b_{t-1}+\textbf x_ty_t^*
\end{align}
In order to avoid calculating the inverse of $\textbf{A}_t$, let $\textbf P_t = (\textbf A_t)^{-1}$.
By using Sherman-Morrison matrix inversion formula \cite{Sherman1950AdjustmentOA}, we can easily get
\begin{align}
\textbf{P}_t = \frac{1}{\lambda}\textbf{P}_{t-1} - \frac{1}{\lambda}\textbf{g}_t(\textbf{u}_t)^\textrm{T}
\end{align}
where $\textbf{u}_t$ and $\textbf{g}_t$ are defined as
\begin{align}
\textbf{u}_t = \textbf{P}_{t-1}\textbf{x}_t
\end{align}
\begin{align}
\textbf{g}_t = \frac{\textbf{u}_t}{\lambda+\textbf{u}_t^\textrm{T}\textbf{x}_t}
\end{align}
where $\textbf{g}_t$ is the gain vector. Plugging (5), (6) into (3), we can finally get the update formula of the weight vector as
\begin{align}
\textbf w_t = \textbf w_{t-1} - \textbf{g}_te_t
\end{align}
where $e_t$ is defined as
\begin{align}
e_t = \textbf{w}^\textrm{T}_{t-1}\textbf x_t -y_t^*
\end{align}

%It can be seen from (4) that the matrix $\textbf{A}_{t}$ is an input autocorrelation symmetric matrix,
%The sum of all elements in its $i^{th}$ column (or row) represents the correlation
%between its $i^{th}$ input feature and other input features. The smaller the sum is, the lower the correlation is.
%Therefore, for the inverse input autocorrelation matrix $\textbf{P}_t$, we can draw the following conclusion:
%the smaller the sum of its $i^{th}$ column (or row) is, the higher the correlation is.

\subsection{CNNs with RLS Optimization}
\subsubsection{Forward Propagation}
\begin{figure}[h]
\centering
\includegraphics[width=0.5\linewidth]{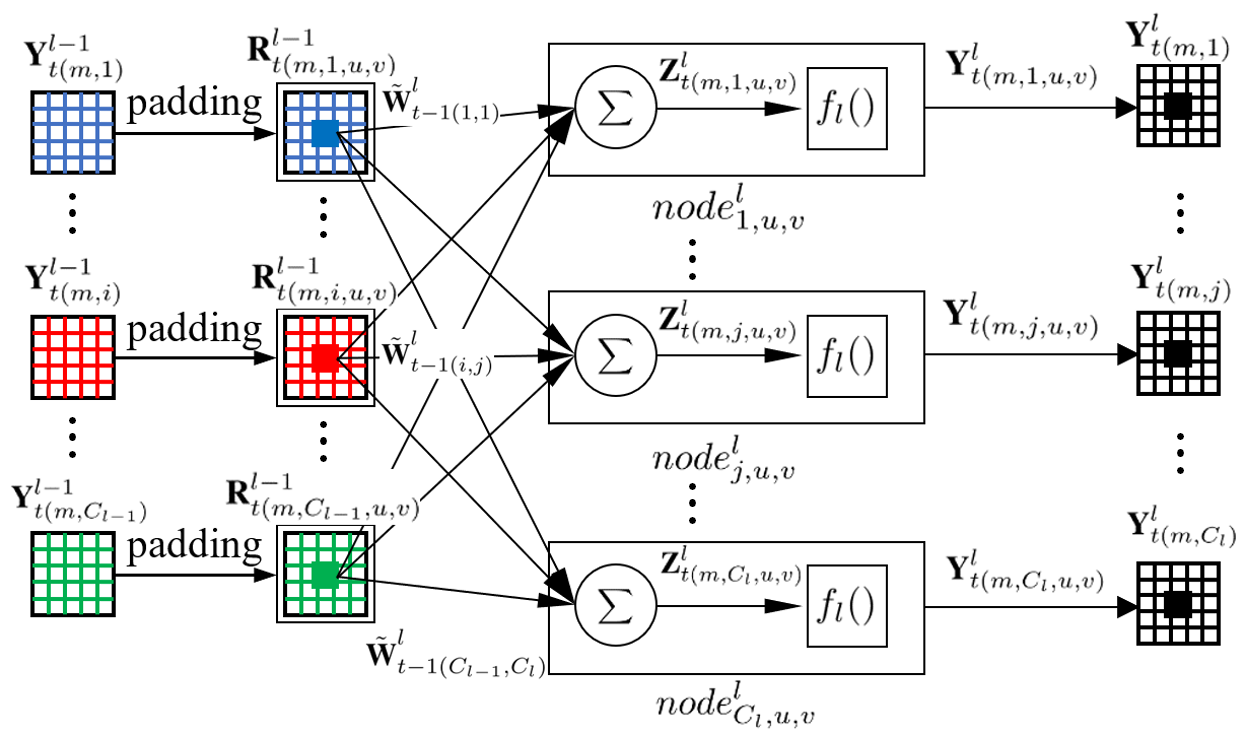}
\caption{Structure of a convolutional layer in a CNN.} %图片标题
\end{figure}

A CNN consists of an input layer followed by several convolutional layers, pooling layers and fully-connected layers.
Since pooling layers have no learnable weights, here we only consider convolutional layers and fully-connected layers.
Let $\textbf{Y}_{t}^{l-1} \in \mathbb{R}^{M\times C_{l-1}\times U_{l-1}\times V_{l-1}}$ and $\textbf{\~W}_{t-1}^{l} \in \mathbb{R}^{C_{l-1}\times C_{l} \times H_l \times W_l}$ denote the output tensor and the filter tensor in a convolutional layer at step $t$,
where $C_l$, $U_l$ and $V_l$ are the number, height and width of output channels, $H_l$ and $W_l$ are the filter height and width, and $M$ is the minibatch size. The structure of a convolutional layer can be illustrated in Figure 1 \cite{Zhang}, where $\textbf{R}_{t}^{l-1} \in \mathbb{R}^{M \times C_{l-1}\times U_{l} \times V_{l} \times H_l \times W_l}$ denotes the receptive fields of all feature maps.
Further, define
\begin{align}
\textbf{X}_{t(m,:,u,v)}^{l} = flatten(\textbf{R}_{t(:,:,u,v,:,:)}^{l-1})
\end{align}
\begin{align}
\textbf{{W}}_{t-1(:,j)}^{l}=flatten(\textbf{\~W}_{t-1(:,j,:,:)}^{l})
\end{align}
where $flatten(\cdot)$ denotes reshaping the given matrix or tensor into a column vector.
Then, the actual output of this convolutional layer can be calculated as
\begin{align}
\textbf{Y}_{t(:,:,u,v)}^{l} = f_{l}(\textbf{X}_{t(:,:,u,v)}^{l}\textbf{{W}}_{t-1}^{l})
\end{align}
where $f_l(\cdot)$ denotes the activation function of this layer.

For a fully-connected layer, the actual activation outputs can be calculated as
\begin{align}
\textbf{Y}_{t}^{l}=f_l(\textbf{X}_{t}^{l}\mathbf{W}_{t-1}^{l})
\end{align}
where $\textbf{X}_{t}^l = \textbf{Y}_{t}^{l-1}$ and $\textbf{W}_{t-1}^{l}$ are the input and  the weight matrix of this layer. For brevity, we omit the bias term in (13) and (14).

\subsubsection{Backward Propagation}
CNNs often use the backward propagation to update  $\{\textbf{W}_{t-1}^l\}_{l=1}^L$ with some optimization algorithms. Zhang et al. propose an RLS optimization algorithm \cite{Zhang}, which can be viewed as a special SGD with the inverse input autocorrelation matrix as the learning rate.
Compared with the conventional first-order optimization algorithm such as SGD and Adam,  it has faster convergence. Therefore, we only review it here.

Let $\textbf{Y}_{t}^{*}$ denote the desired output of the current minibatch input $\textbf{Y}_{t}^{0}$, and define a Mean Squared Error (MSE) loss function as
\begin{align}
J(\textbf{W}_{t-1})=\frac{1}{2M} \left\|\textbf{Z}_{t}^{L}-\textbf{Z}_{t}^{L*}\right\|_\textsc{F}^2
\end{align}
where $\textbf{W}_{t-1}$ denotes all weights in the CNN at step $t$,
$\mathbf{Z}_{t}^{L*}=f_L^{-1}(\textbf{Y}_{t}^*)$ and $\mathbf{Z}_{t}^{L} =f_L^{-1}(\textbf{Y}_{t}^L)$.
Then, by using the RLS optimization \cite{Zhang}, the recursive update rule of $\textbf{P}_t^l = (\textbf{A}_t^l)^{-1}$ is defined as
\begin{align}
\textbf{P}_t^l \approx \frac{1}{\lambda} \textbf{P}_{t-1}^l -  \frac{k}{\lambda h_t^l} \textbf{u}_{t}^l (\textbf{u}_{t}^l)^{\textrm{T}}
\end{align}
where $k>0$ is the average scaling factor. Here,  $\textbf{u}_t^l$ and $h_t^l$ are defined as follows
\begin{align}
\textbf{u}_t^l=\textbf{P}_{t-1}^l {{\textbf x}}_{t}^l ~~~~ \\
h_{t}^l=\lambda+k({{\textbf x}}_{t}^l)^{\textrm{T}}\textbf{u}_t^l
\end{align}
where ${{\textbf x}}_{t}^l$ is the average vector. For a fully-connected layer, ${{\textbf x}}_{t}^l$ is defined as
\begin{align}
{{\textbf x}}_{t}^l=\frac{1}{M} \sum_{m=1}^M (\textbf{X}_{t(m,:)}^l)^{\textrm{T}}
\end{align}
For a convolutional layer, $\textbf{{x}}_{t}^l$ is defined as
\begin{align}
{{{\textbf x}}}_{t}^l=\frac{1}{M U_l V_l} \sum_{m=1}^M \sum_{u=1}^{U_l} \sum_{v=1}^{V_l} (\textbf{X}_{t(m,:,u,v)}^l)^{\textrm{T}}
\end{align}
The recursive update rule of $\textbf{W}_t^l$ is defined as
\begin{align}
\mathbf{\Psi}_t^l = \alpha\mathbf{\Psi}_{t-1}^l -  \frac{\eta^l}{ h_t^l} \textbf{P}_{t-1}^l {\mathbf{\nabla}}_{\textbf{W}_{t-1}^{l}}
\end{align}
\begin{align}
\textbf{W}_t^l \approx \textbf{W}_{t-1}^l + \mathbf{\Psi}_t^l
\end{align}
where $\mathbf{\Psi}_t^l$ is the velocity matrix of the $l^{th}$ layer at step $t$, $\alpha$ is the momentum factor, $\eta^l > 0$ is the gradient scaling factor, and $\mathbf{\nabla}_{\textbf{W}_{t-1}^{l}}$ denotes $\partial{J}(\textbf{W}_{t-1})/\partial \textbf{W}^{l}_{t-1}$.

\section{The Proposed Algorithm}
In this section, we first introduce the theoretical foundation of our proposed algorithm. Then,
by combining inverse autocorrelation matrices and weight matrices to evaluate and prune unimportant input channels or nodes layer by layer,
we propose an RLS-based training and pruning algorithm for CNNs.

\subsection{Theoretical Foundation}
The general goal of a pruning algorithm is to delete redundant and less-informative channels and nodes in CNNs.
As mentioned in Section 1, existing pruning algorithms generally use the weight change or the accuracy change to evaluate the importance of channels and  nodes in CNNs, but don't consider the importance of input components. However,
From (14) and (15), the output of each layer is determined by its input  and weight matrix.
In this subsection, we show that we can use $\textbf{P}_t$  to measure the importance of input components.

As introduced in Section 2.1, $\textbf{A}_t$ is the input autocorrelation matrix. Let the input feature $\textbf x_t=[x_{t(1)},x_{t(2)},\cdots,x_{t(n)}]^\textrm{T}$ and $s_{\textbf x_t}$ be the sum of $\textbf x_t$. Then, $\textbf x_t \textbf x_t^\textrm{T}$ is calculated as
\begin{align}
\textbf x_t \textbf x_t^\textrm{T} =
\left [
\begin{array}{cccc}
x_{t(1)}x_{t(1)} & x_{t(1)}x_{t(2)} & \cdots & x_{t(1)}x_{t(n)} \\
x_{t(2)}x_{t(1)} & x_{t(2)}x_{t(2)} & \cdots & x_{t(2)}x_{t(n)} \\
\cdots & \cdots & \cdots & \cdots \\
x_{t(n)}x_{t(1)} & x_{t(n)}x_{t(2)} & \cdots & x_{t(n)}x_{t(n)} \\
\end{array}
\right ]
\end{align}
The sum of the $i^{th}$ row (or column) of $\textbf A_t$ can be defined as
\begin{align}
s_{\textbf A_{t(i)}} = \sum_{k=1}^t\lambda^{t-k} x_{k(i)} s_{\textbf x_k}=\lambda s_{\textbf A_{t-1(i)}} + x_{t(i)} s_{\textbf x_t}
\end{align}
which means $s_{\textbf A_{t(i)}}$ is approximately proportional to the $i^{th}$ component of all inputs.
If $s_{\textbf A_{t(i)}}$ is small, the $i^{th}$ component of all inputs  will probably be small,
and its influence on the output will probably be small. Since $\textbf{P}_t$ is the inverse of $\textbf{A}_t$, we can easily draw a conclusion:
If the sum of the $i^{th}$ row (or column) of $\textbf{P}_t$ is big, the importance of the $i^{th}$ component of all inputs will probably be small.
Similarly, for fully-connected layers in CNNs, $\textbf{P}_t^l$ can be used to measure the importance of their input nodes.
For convolutional layers in CNNs, existing pruning algorithms generally delete their unimportant channels rather than their nodes. From (11) and (19),
we can get ${\bar{\textbf x}}_{t}^l \in \mathbb{R}^{C_{l-1} H_l W_l}$.  It means that an input channel includes $H_l W_l$ rows (or columns) of $\textbf{A}_t^l$. Similar to (24), the sum of $i$ to $j$ rows (or columns) in $\textbf{A}_t$ is
\begin{align}
s_{\textbf A_{t(i:j)}} = \sum_{k=1}^t\sum_{a=i}^j \lambda^{t-k} x_{k(a)} s_{\textbf x_k}=\lambda s_{\textbf A_{t-1(i:j)}} + \sum_{a=i}^j x_{t(a)} s_{\textbf x_t}
\end{align}
which means $s_{\textbf A_{t(i:j)}}$ is approximately proportional to the sum of the $i^{th}$ to $j^{th}$ components of all inputs.
Thus, for convolutional layers in CNNs, $\textbf{P}_t^l$ can be used to measure the importance of their input channels as well.
\subsection{RLS-Based Pruning}
Firstly, based on the above theoretical foundation, we define a vector $\textbf s_{\textbf P_t^l}$ to represent the importance of input channels or nodes.
The $i^{th}$ element of $\textbf s_{\textbf P_{t}^l}$ is defined as
\begin{equation}
\textbf s_{\textbf P_{t(i)}^l} =
\left\{
\begin{aligned}
\sum_{j=1}^{C_{l-1}H_l W_l}\sum_{g=G_{i-1}}^{G_i-1}\textbf{P}_{{t}(g,j)}^l  ~~~~~~~~~~~~~~~l\leq L_c  \\
\sum_{j=1}^{N_{l-1}}\textbf{P}_{{t}(i,j)}^l                     ~~~~~~~~~~~~~~~~~~~~~~~~~l>L_c
\end{aligned}
\right.
\end{equation}
where $G_i=H_l W_l \times i+1$, $1\leqslant i \leqslant C_{l-1}$,
$L_c$ denotes the total number of convolutional layers, and $N_{l-1}$ denotes the total number of input nodes in the current fully-connected layer.

Secondly, we further use the weight matrix to measure the importance of input channels or nodes.
Li et al. \cite{DBLP:conf/iclr/0022KDSG17} have demonstrated that the $L_1$-norm of weight matrices can be used to measure
the importance of output features.
From (12), (13) and Figure 1, $\textbf{\~W}_{t-1}^{l}$ will influence the output of the $l^{th}$ layer (i.e., the input of the $(l+1)^{th}$ layer),
so  we can use  $\textbf{\~W}_{t-1}^{l-1}$ to evaluate the importance of the input channel in $l^{th}$ layer.
Let $\textbf s_{\textbf{W}_{t}^{l}}$ denote the $L_1$-norm of $\textbf{\~W}_{t-1}^{l-1}$.
The $i^{th}$ element of $\textbf s_{\textbf{W}_{t}^{l}}$ is defined as
\begin{equation}
\textbf s_{\textbf{W}_{t(i)}^{l}} =
\left\{
\begin{aligned}
\sum_{j=1}^{C_{l-2}}\sum_{u=1}^{H_{l-1}}\sum_{v=1}^{W_{l-1}} \big|\textbf{\~W}_{t-1(j,i,u,v)}^{l-1}\big|~~~~~~~~~~~~~2\leq l\leq L_c \\
\sum_{j=1}^{N_{l-1}} \big|\textbf{\~W}_{t-1(j,i)}^{l-1}\big|~~~~~~~~~~~~~~~~~~~~~~~~~~~~ l>L_c ~~~~
\end{aligned}
\right.
\end{equation}
where $| \cdot |$ denotes the absolute value of a real number.

Thirdly, we perform sort operation on $\textbf s_{\textbf P_{t}^l}$ and $\textbf s_{\textbf W_{t}^l}$, namely
\begin{align}
\textbf k_{t(\textbf P)}^l = \texttt{Sort}_{\texttt{d}}(\textbf s_{\textbf P_{t}^l})
\end{align}
\begin{align}
\textbf k_{t(\textbf W)}^l = \texttt{Sort}_{\texttt{a}}(\textbf s_{\textbf W_{t}^l})
\end{align}
where $\texttt{Sort}_{\texttt{d}}(\cdot)$ and $\texttt{Sort}_{\texttt{a}}(\cdot)$ denote sorting in descending and ascending order, and $\textbf k_{t(\textbf P)}^l$ and $\textbf k_{t(\textbf W)}^l$ are the sorted index sets, respectively.

Then, we can define an intersection set $\textbf s_{\textbf D_{t}^l}$ for pruning, that is,
\begin{equation}
\textbf s_{\textbf D_{t}^l} =
\left\{
\begin{aligned}
\texttt{Cut}_\xi (\textbf k_{t(\textbf P)}^l)~ \cap ~\texttt{Cut}_\xi (\textbf k_{t(\textbf W)}^{l})~~~~~~~~~~~~ l\neq1 \\
\texttt{Cut}_{0.5\xi} (\textbf k_{t(\textbf P)}^l)~~~~~~~~~~~~~~~~~~~~~~~~~~~~~~~~~ l=1
\end{aligned}
\right.
\end{equation}
where $\xi$ is the pruning ratio, and $\texttt{Cut}_\xi(\cdot)$ denotes the subset composed of the top $\xi$ elements in the given set.
For the input layer, since $\textbf s_{\textbf{W}_{t}^{l}}$ and $\textbf k_{t(\textbf W)}^l$ do not exist,
we only use $\textbf k_{t(\textbf P)}^l$ for pruning. In addition, the input nodes of this layer are in fact the original input features,
so we suggest a small pruning ratio such as $0.5\xi$ for avoiding overpruning. In particular, for the fully-connected layer adjacent to the convolutional layer, pruning an input node will make the corresponding output channel in the preceding layer become incomplete. For this case,  we choose to delete the corresponding output channel.

Finally, we determine when to prune the network.
As mentioned in Section 1, existing algorithms generally perform the one-shot pruning, which cannot automatically adapt to different scale CNNs.
If the pruning ratio is set to too high, the performance of the pruned CNN won't recover.
Therefore, in order to prune CNNs to a reasonable level, we choose to prune the network many times.
After training some steps, we will record the current loss and do the first pruning of the network.
Then, our algorithm will continue to train the pruned network, and won't do the next pruning until the loss is fully reduced.

In conclusion, the RLS algorithm for training and pruning CNNs can be summarized as Algorithm 1, where $q$ denotes the basic training epochs, respectively.

\begin{algorithm}[h]
\caption{RLS for Training and Pruning CNNs }
\label{alg:algorithm}
\textbf{Input}: {$M$, $\lambda$, $k$, $\alpha$, $\{\eta^l\}_{l=1}^{L}$}, $\xi$, $q$ \\
\textbf{Initial}: $\{\textbf{W}_0^l\}_{l=1}^{L}$, $\{\textbf{P}_0^l \}_{l=1}^L$, $t = 1$, $loss=10^5$\\
\vspace{-0.4cm}
\begin{algorithmic}[1] %[1] enables line numbers
\WHILE{stopping criterion not met}
\STATE Get the minibatch $(\textbf{Y}_t^0,\textbf{Y}_t^*)$ from the training dataset $\mathcal{D}$
\FOR {$l=1$ to $L$}
\STATE  Compute the outputs by (13) or (14);
\ENDFOR \STATE Compute $J(\textbf{W}_{t-1})$ by (15)
\FOR {$l=L$ to $1$}
\STATE Compute ${\nabla}_{\textbf{W}_{t-1}^l}$;
\STATE Update $\mathbf{\Psi}_{t-1}^l$, $\textbf{W}_{t-1}^l$, $\textbf{P}_{t-1}^{l}$ by (21), (22) and (16)
\ENDFOR
\IF { $t> \frac{q|\mathcal{D}|}{M}$ and $tM \% |\mathcal{D}| ==0$ and $J(\textbf{W}_{t-1})<loss$}
\STATE Update  $loss = J(\textbf{W}_{t-1}) $
\FOR {$l=1$ to $L$}
\STATE Compute $\textbf s_{\textbf P_{t}^l}$, $\textbf s_{\textbf W_{t}^{l}}$ by (26), (27)
\STATE Get $\textbf k_{t(\textbf P)}^l$, $\textbf k_{t(\textbf W)}^l$ by (28), (29)
\STATE Get $\textbf s_{\textbf D_{t}^l}$ by (30)
\STATE Prune input channels or nodes indexed in $\textbf s_{\textbf D_{t}^l}$
\ENDFOR
\ENDIF
\STATE Update $t = t+1$
\ENDWHILE
\end{algorithmic}
\end{algorithm}

\section{Discussion}
In this section, we briefly introduce the related network pruning algorithms proposed in recent years,
and outline the advantages of our proposed algorithm.

Existing network pruning algorithms are generally based on two mainstream views.
First, people think it is important to start with training a large, over-parameterized network \cite{DBLP:conf/cvpr/Carreira-Perpinan18}.
Because of its stronger representation and optimization capabilities, a set of redundant weights can be safely deleted from it without significantly compromising accuracy. It is generally considered that this is better than training a smaller network directly from scratch \cite{DBLP:conf/cvpr/Yu00LMHGLD18}.
Second, it is believed that the pruned architecture and its associated weights are essential to obtain the final effective model \cite{DBLP:conf/nips/HanPTD15}.
Therefore, existing network pruning algorithms choose to fine-tune the pruned model instead of training from scratch.
Usually, network pruning can be classified into unstructured pruning and structured pruning.

Unstructured pruning sets some weights or nodes to zero for simulating deletion.
Unstructured pruning can be traced back to Optimal Brain Damage \cite{DBLP:conf/nips/CunDS89}, which prunes the weights based on the Hessian matrix of the loss function. Molchanov et al. use variational dropout \cite{DBLP:conf/nips/BlumHP15} to prune redundant weights \cite{DBLP:conf/icml/MolchanovAV17}.
Louizos et al. learn sparse networks through $L_0$ norm regularization based on random gates \cite{DBLP:journals/corr/abs-1712-01312}. Kang and Han mask negative values based on batch normalization and rectified linear units \cite{DBLP:conf/icml/KangH20}.
Unstructured pruning does not substantially delete filters or nodes, so it will not have an irreversible impact on networks. However, this also results in compression and acceleration that can't be achieved without dedicated hardware \cite{DBLP:conf/cvpr/HeZRS16}.

In contrast, structured pruning deletes nodes or channels or even the entire layer directly.
Among structured pruning algorithms, channel pruning is most popular because it runs at the finest-grained level and is suitable for traditional deep learning frameworks. Since the structured pruning may have a difficult or even unrecoverable impact on networks, how to find a suitable set of pruning channels has become the most important research direction.
Li et al. and Hu et al. respectively trim the channel by using filter weight norm and the average percentage of zeros in the output \cite{DBLP:conf/iclr/0022KDSG17,DBLP:journals/corr/HuPTT16}.
Group sparsity is also widely used in smooth pruning after training.
Liu et al. impose sparsity constraints on channel scaling factors during training \cite{DBLP:journals/corr/abs-1708-06519}, and then use it for channel pruning.
Ma et al. prune the channel through sparse convolution mode and group sparsity \cite{DBLP:conf/aaai/MaGNL0MRW20}.

Our algorithm belongs to structured pruning. Compared with existing algorithms, our algorithm has the following new properties.
1) By using the RLS optimization \cite{Zhang},  it has faster convergence and higher precision.
2) It combines both input features and weight matrices to measure the importance of input channels and nodes, so it has the higher pruned precision. In additon, It can prune the originalfeatures of samples.
3) It combines pruning and training. Once training is completed, pruning is also completed at the same time, which makes the pruning more efficient.
4) It performs the multi-shot pruning instead of the one-shot pruning. Each new pruning doesn't be performed until the pruned network recovers to the original performance, so it can prune different scale networks adaptively and prevent the pruning loss from becoming too large.

\section{Experiments}
In this section, we demonstrate the effectiveness of our algorithm by three experiments, pruning FNNs on MNIST \cite{726791}, pruning CNNs on  CIFAR-10 \cite{Krizhevsky2009LearningML}  and  pruning CNNs on SVHN \cite{Netzer2011ReadingDI}. We choose four popular pruning algorithms, $L_1$-norm \cite{DBLP:conf/iclr/0022KDSG17}, Net Slimming \cite{DBLP:journals/corr/abs-1708-06519}, Soft Pruning \cite{DBLP:conf/ijcai/HeKDFY18} and Weight Level \cite{DBLP:conf/nips/HanPTD15}, for comparsion.

\subsection{Pruning FNNs on MNIST}
In this experiment, we will use our algorithm to train and prune FNNs on the MNIST dataset for verifying its performance. In additon, we will
evaluate the effect of $\xi$ on its performance.

All tested algorithms run 200 epochs on one 2060 GPU. Their FNNs all consist of four layers. The first layer is the input layer for 28$\times$28 images. The second and third layers are fully-connected ReLU layers, which respectively have 1024 and 512 nodes. The fourth layer is a fully-connected linear output layer with 10 nodes. All network weights are initialized with the default settings of PyTorch, and the minibatch size $M$ is 128.
In addition, in our algorithm, $\alpha$, $\lambda$, $\xi$, $k$, $\{\eta^l\}_{l=1}^{L}$,  and $q$ are set to 0.5, 1, 40\%, 0.1, ${1}$ and $30$, respectively.
The four compared algorithms all use the Momentum optimization with the learning rate 0.1, the momentum factor 0.9 and the weight decay 0.0001.
Their pruning ratios for each hidden layer are set as follows:
1) $L_1$-norm prunes 70\%  nodes with the smallest $L_1$-norm.
2) Net Slimming adds a batch normalization layer after each hidden layer, uses $L_1$ regularization for the scaling factors of batch normalization layers, and then prunes 70\% nodes with the smallest scaling factors.
3) Soft Pruning zeroes out the smallest 30\% absolute values of weights at each epoch.
4) Weight Level uses $L_2$ regularization to evaluate the importance of weights, and then zeroes out 70\% weights with the lowest importance.

The comparison result of all algorithms for pruning FNNs on MNIST is shown in Table 1, where
\emph{n} and \emph{w} denote the total number of nodes and weights of each layer in the original network,
\emph{n}(\%) and \emph{w}(\%) denote the percentage of retained nodes and weights,
and the last two lines show the average classifiaction precision and MSE loss of the last 10 epochs by using unpruned and pruned FNNs.
From Table 1, it is clear that the unpruned FNN with our algorithm has the highest precision and smallest loss,
which indicates that the RLS optimization can achieve better performance than the Momentum optimization.
More importantly, it shows that our pruned FNN  retains 41.1\% nodes and 16.4\% weights. It is obviously superior to
other four pruned FNNs. In particular, our algorithm also prunes 67\% input nodes, which inicates that our algorithm
can select the original input features automatically.
Some readers maybe argue that our pruned precision and loss are inferior to those of Soft Pruning and Weight Level.
This is because both of them are unstructured pruning algorithms. As mentioned in Section 4, they only zero out some weights of each hidden
layer but don't really prune these weights. In addition, they retain a much higher proportion of weights than our algorithm.
Comparing with two pruned networks using $L_1$-norm and Net Slimming structured pruning algorithms, our pruned network has
higher precision and smaller loss under lesser retained nodes and weights.

Figure 2 shows the effect of $\xi \in \{20\%, 30\%, 40\%, 50\%\}$ on our algorithm for pruning FNNs on MNIST.
In Figure 2, three subfigure show the percentage variation of retained nodes, retained weights and classification precision, respectively.
It is clear that our algorithms with four different pruning ratios all perform multiple pruning.
A larger $\xi$ will result in more nodes and weights pruned from FNNs.
But as $\xi$ and training epochs get larger, the pruning becomes less frequent, since our algorithm
don't perform a new pruning unless the MSE loss of pruned networks has reduced to the orginal level.
From Figure 2(c), final classification precision is affected by the selection of $\xi$ slightly.

\begin{table}[ht]
  \centering
  \caption{Comparison result of five algorithms for pruning FNNs on MNIST.}
  \renewcommand\arraystretch{1.3}
  \fontsize{7.5pt}{8pt}\selectfont
  \label{tab:performance_comparison}
    \begin{tabular}{|c|c|c|p{0.8cm}<{\centering}|p{0.8cm}<{\centering}|
    p{0.8cm}<{\centering}|p{0.8cm}<{\centering}|p{0.8cm}<{\centering}|p{0.8cm}<{\centering}|
    p{0.8cm}<{\centering}|p{0.8cm}<{\centering}|p{0.8cm}<{\centering}|p{0.8cm}<{\centering}|}
    \hline
    \multirow{2}{*}{Layer}&
    \multicolumn{2}{c|}{FNNs}&\multicolumn{2}{c|}{$L_1$-norm}&\multicolumn{2}{c|}{Net Slimming}&\multicolumn{2}{c|}{Soft Pruning}&\multicolumn{2}{c|}{Weight Level}&\multicolumn{2}{c|}{RLS Pruning}\cr\cline{2-13}
    &\emph{n} &\emph{w} &\emph{n}{(\%)}&\emph{w}{(\%)}&\emph{n}{(\%)}&\emph{w}{(\%)}&\emph{n}{(\%)}&\emph{w}{(\%)}&\emph{n}{(\%)}&\emph{w}{(\%)}&\emph{n}{(\%)}&\emph{w}{(\%)}\cr
    \hline
    input &784 &-      &100.0&-   &100.0&-   &100.0&-   &100.0&-   &33.0 &-   \cr\hline
    fc1   &1024&802816 &30.0 &30.0&35.9 &35.9&100.0&70.0&100.0&32.0&37.1 &12.3\cr\hline
    fc2   &~~~~512~~~~ &524288 &30.0 &9.0 &16.6 &6.0 &100.0&70.0&100.0&26.3&60.0 &22.2\cr\hline
    fc3   &10  &5120   &100.0&30.0&100.0&16.6&100.0&70.0&100.0&74.8&100.0&60.0\cr\hline
    Total &2330&1332224&53.8 &21.7&53.4 &24.1&100.0&70.0&100.0&30.0&41.1 &16.4\cr\hline
    \multicolumn{3}{|c|}{(Unpruned Prec~,~Pruned Prec)}&
    \multicolumn{2}{c|}{(98.9~,~98.4)}&
    \multicolumn{2}{c|}{(98.8~,~98.4)}&
    \multicolumn{2}{c|}{(98.9~,~98.8)}&
    \multicolumn{2}{c|}{(98.8~,~98.8)}&
    \multicolumn{2}{c|}{(99.3~,~98.5)}\cr\hline

    \multicolumn{3}{|c|}{(Unpruned Loss~,~Pruned Loss)}&
    \multicolumn{2}{c|}{(0.016~,~0.020)}&
    \multicolumn{2}{c|}{(0.024~,~0.027)}&
    \multicolumn{2}{c|}{(0.016~,~0.017)}&
    \multicolumn{2}{c|}{(0.016~,~0.016)}&
    \multicolumn{2}{c|}{(0.011~,~0.020)}\cr
    \hline
    \end{tabular}
\end{table}

\begin{figure*}[ht]
\centering
\includegraphics[width=1.02\textwidth]{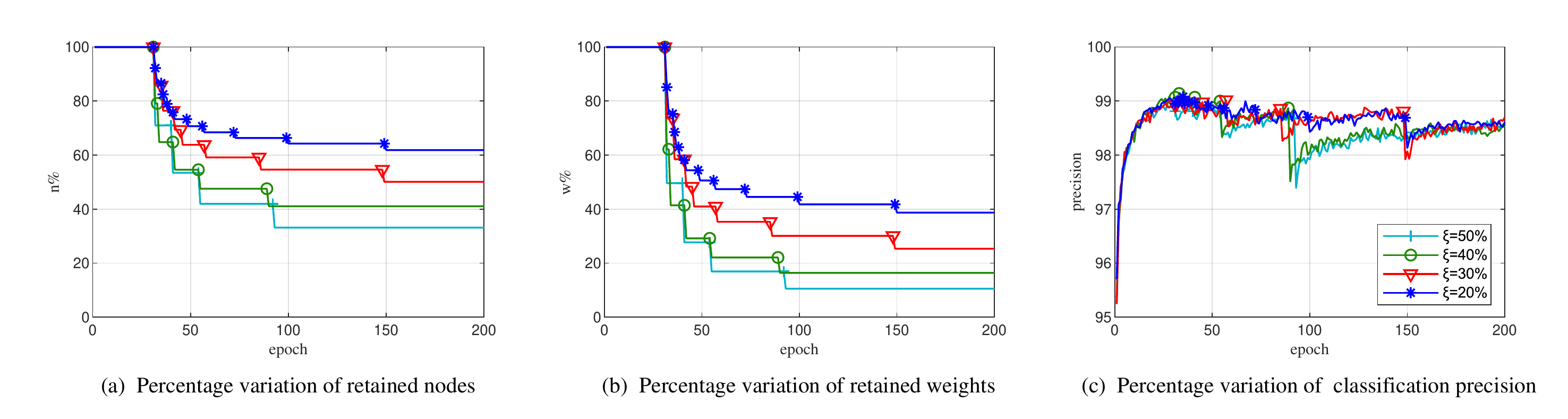}
\caption{Effect of different $\xi$ on our algorithm for pruning FNNs on MNIST.} %图片标题
\end{figure*}

\subsection{Pruning CNNs on CIFAR-10}
In this experiment, we will use our algorithm to train and prune CNNs on the CIFAR-10 dataset for verifying its performance. In additon, we will
evaluate the effect of $\xi$ on its performance.

Similar to in the first experiment, all tested algorithms run 200 epochs. Their CNNs are a mini-VGG network, which consist of eleven layers. The first layer is the input layer for $32 \times 32 \times 3$ images. The $2^{th}$, $3^{th}$, $5^{th}$, $6^{th}$ and $8^{th}$ layers are convolution ReLU layers, which respectively have 64, 64, 128, 128 and 256 output channels. All convolution layers use $3 \times 3$ kernel sizes and 1 stride.
The $10^{th}$ layer is a fully-connected ReLU layer with 1024 output nodes. The last layer is a fully-connected linear output layer with 10 nodes.
The other layers are  maxpool layers.
The pruning ratios of four compared algorithms for each hidden layer are set as follows:
1) $L_1$-norm and Net Slimming prune 50\% convolutional channels  and 50\% fully-connected nodes.
2) Soft Pruning zeroes out 30\% weights and Weight Level zeroes out 50\% weights.
All other settings for five algorithms are the same as those in the first experiment.
Notice that our algorithm won't prune the input layer since it only has three input channels in this experiment.

The comparison result of all algorithms for pruning CNNs on CIFAR-10 is shown in Table 2.
It shows that our unpruned CNN has the best performance than four compared unpruned CNNs,
and Net Slimming prunes the most weights and $L_1$-norm prunes the most nodes.
However, we can see that the pruning of Net Slimming is uneven.
It prunes few output channels in conv2 and conv3 layers but almost all output channels in conv5 layer,
which result in the lowest precision and the biggest loss.
Our algorithm prunes the second most weights and nodes, but its
unpruned and pruned CNNs have the higher precision and smaller loss than those with
Net Slimming and $L_1$-norm. Among five pruned CNNs, the CNN with Weight Level recovers its performance
and achieve the highest precesion and the smallest loss. Whereas, as mentioned in the first experiment,
unstructured pruning algorithms don't really prune any nodes and weights.

Figure 3 shows the effect of $\xi \in \{20\%, 30\%, 40\%, 50\%\}$ on our  algorithm for pruning CNNs on CIFAR-10.
The conclusions are similar to those in the first experiment.

\begin{table*}[ht]
  \centering
  \caption{Comparison result of five algorithms for pruning CNNs on CIFAR-10.}
  \renewcommand\arraystretch{1.3}
  \fontsize{7.5pt}{8pt}\selectfont
  \label{tab:performance_comparison}
    \begin{tabular}{|c|c|c|p{0.8cm}<{\centering}|p{0.8cm}<{\centering}|
    p{0.8cm}<{\centering}|p{0.8cm}<{\centering}|p{0.8cm}<{\centering}|p{0.8cm}<{\centering}|
    p{0.8cm}<{\centering}|p{0.8cm}<{\centering}|p{0.8cm}<{\centering}|p{0.8cm}<{\centering}|}
    \hline
    \multirow{2}{*}{Layer}&
    \multicolumn{2}{c|}{CNNs}&\multicolumn{2}{c|}{$L_1$-norm}&\multicolumn{2}{c|}{Net Slimming}&\multicolumn{2}{c|}{Soft Pruning}&\multicolumn{2}{c|}{Weight Level}&\multicolumn{2}{c|}{RLS Pruning}\cr\cline{2-13}
    &n&w&n\%&w\%&n\%&w\%&n\%&w\%&n\%&w\%&n\%&w\%\cr
    \hline
    input &3072   &-      &100.0&-   &100.0&-   &100.0&-   &100.0&-   &100.0 &-   \cr\hline
    conv1 &65536  &1728   &50.0 &50.0&32.8 &32.8&100.0&70.0&100.0&90.0&62.5  &62.5\cr\hline
    conv2 &65536  &36864  &50.0 &25.0&95.3 &31.3&100.0&70.0&100.0&81.4&46.9  &29.3\cr\hline
    conv3 &32768  &73728  &50.0 &25.0&91.4 &87.1&100.0&70.0&100.0&93.6&48.4  &22.7\cr\hline
    conv4 &32768  &147456 &50.0 &25.0&78.9 &72.1&100.0&70.0&100.0&92.8&50.0  &24.2\cr\hline
    conv5 &16384  &294912 &50.0 &25.0&7.4  &5.9 &100.0&70.0&100.0&91.5&25.0  &12.5 \cr\hline
    fc1   &1024   &4194304&50.0 &25.0&49.4 &3.7 &100.0&70.0&100.0&44.4&68.3  &17.1 \cr\hline
    fc2   &10     &10240  &100.0&50.0&100.0&49.4&100.0&70.0&100.0&84.1&100.0 &68.3\cr\hline
    Total &~217098~ &4759232&50.7 &25.1&66.6 &7.5 &100.0&70.0&100.0&50.0&51.5  &17.3\cr\hline
    \multicolumn{3}{|c|}{(Unpruned Prec~,~Pruned Prec)}&
    \multicolumn{2}{c|}{(89.7~,~86.3)}&
    \multicolumn{2}{c|}{(88.1~,~84.9)}&
    \multicolumn{2}{c|}{(89.7~,~84.7)}&
    \multicolumn{2}{c|}{(89.9~,~89.9)}&
    \multicolumn{2}{c|}{(91.4~,~88.3)}\cr\hline

    \multicolumn{3}{|c|}{(Unpruned Loss~,~Pruned Loss)}&
    \multicolumn{2}{c|}{(0.084~,~0.108)}&
    \multicolumn{2}{c|}{(0.103~,~0.124)}&
    \multicolumn{2}{c|}{(0.083~,~0.123)}&
    \multicolumn{2}{c|}{(0.083~,~0.084)}&
    \multicolumn{2}{c|}{(0.070~,~0.101)}\cr
    \hline
    \end{tabular}
\end{table*}
\begin{figure*}[ht]
\centering
\includegraphics[width=1.02\textwidth]{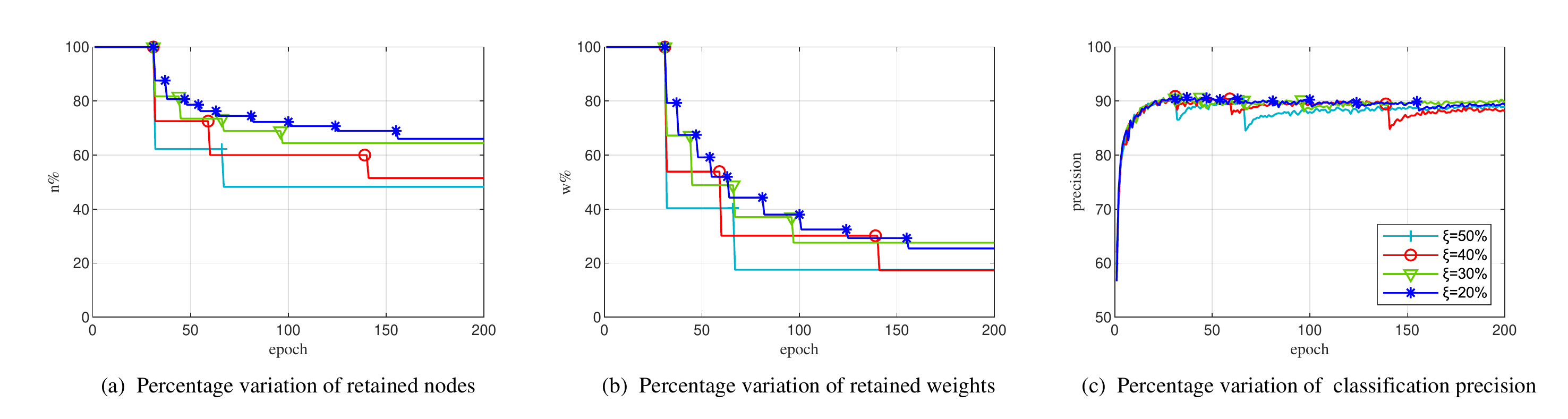}
\caption{Effect of different $\xi$ on our algorithm for pruning CNNs on CIFAR-10.} %图片标题
\end{figure*}

\subsection{Pruning CNNs on SVHN}
To further validate the effectiveness of our algorithm, we train and prune CNNs on the SVHN dataset in this experiment.
All network and algorithm settings are the same as those in the second experiment.

Table 3 shows the comparision result of five algorithms for pruning CNNs on SVHN.
It shows that our unpruned CNN still has the best performance than four compared unpruned CNNs.
In addition, our pruned CNN  prunes the most weights and nodes, and its precision and loss are
only slights worse than those of our unpruned CNN. Pruned CNNs with Soft Pruning and Weight Level
have amazing performances. The reason has been given in the first experiment. They seems impractical
since they can't really compress CNNs.
Figure 4 shows the effect of $\xi \in \{20\%, 30\%, 40\%, 50\%\}$ on our algorithm for pruning CNNs on SVHN.
The conclusions are also similar to those in the first experiment.
By comparing Figure 4 and Figure 3, although the original CNN for SVHN is the same as the that for CIFAR-10,
the pruning percentage and recovery speed in this experiment are obviously higher than those in the second
experiment, since the SVHN task is easier than the CIFAR-10 task.
Further, by comparing Table 3 and Table 2, the pruning percentages of four compared algorithm are almost unchaged
in both experiments. It demonstrates that our algorithm can prune CNNs to a reasonable level
for fitting the task's difficulty level.

\begin{table*}[ht]
  \centering
  \renewcommand\arraystretch{1.2}
  \fontsize{7.5pt}{8pt}\selectfont
  \caption{Comparison result of five algorithms for pruning CNNs on SVHN.}
  \renewcommand\arraystretch{1.3}
  \fontsize{7.5pt}{8pt}\selectfont
  \label{tab:performance_comparison}
    \begin{tabular}{|c|c|c|p{0.8cm}<{\centering}|p{0.8cm}<{\centering}|
    p{0.8cm}<{\centering}|p{0.8cm}<{\centering}|p{0.8cm}<{\centering}|p{0.8cm}<{\centering}|
    p{0.8cm}<{\centering}|p{0.8cm}<{\centering}|p{0.8cm}<{\centering}|p{0.8cm}<{\centering}|}
    \hline
    \multirow{2}{*}{Layer}&
    \multicolumn{2}{c|}{CNNs}&\multicolumn{2}{c|}{$L_1$-norm}&\multicolumn{2}{c|}{Net Slimming}&\multicolumn{2}{c|}{Soft Pruning}&\multicolumn{2}{c|}{Weight Level}&\multicolumn{2}{c|}{RLS Pruning}\cr\cline{2-13}
    &n&w&n\%&w\%&n\%&w\%&n\%&w\%&n\%&w\%&n\%&w\%\cr
    \hline
    input &3072   &-      &100.0&-   &100.0&-   &100.0&-   &100.0&-   &100.0 &-   \cr\hline
    conv1 &65536  &1728   &50.0 &50.0&45.3 &45.3&100.0&70.0&100.0&92.0&45.3  &45.3\cr\hline
    conv2 &65536  &36864  &50.0 &25.0&93.8 &42.5&100.0&70.0&100.0&74.6&37.5  &17.0\cr\hline
    conv3 &32768  &73728  &50.0 &25.0&78.9 &74.0&100.0&70.0&100.0&87.7&35.9  &13.5\cr\hline
    conv4 &32768  &147456 &50.0 &25.0&68.0 &53.6&100.0&70.0&100.0&95.6&39.8  &14.3\cr\hline
    conv5 &16384  &294912 &50.0 &25.0&16.4 &11.2&100.0&70.0&100.0&97.0&17.2  &6.8\cr\hline
    fc1   &1024   &4194304&50.0 &25.0&49.4 &8.1 &100.0&70.0&100.0&44.1&34.2  &5.9 \cr\hline
    fc2   &10     &10240  &100.0&50.0&100.0&49.4&100.0&70.0&100.0&91.8&100.0 &4.2\cr\hline
    Total &~217098~ &4759232&50.7 &25.1&67.0 &11.1&100.0&70.0&100.0&50.0&39.3  &6.5\cr\hline

    \multicolumn{3}{|c|}{(Unpruned Prec~,~Pruned Prec)}&
    \multicolumn{2}{c|}{(94.5~,~93.7)}&
    \multicolumn{2}{c|}{(92.3~,~91.2)}&
    \multicolumn{2}{c|}{(94.5~,~94.4)}&
    \multicolumn{2}{c|}{(94.5~,~94.5)}&
    \multicolumn{2}{c|}{(94.7~,~94.1)}\cr\hline

    \multicolumn{3}{|c|}{(Unpruned Loss~,~Pruned Loss)}&
    \multicolumn{2}{c|}{(0.049~,~0.055)}&
    \multicolumn{2}{c|}{(0.077~,~0.081)}&
    \multicolumn{2}{c|}{(0.049~,~0.050)}&
    \multicolumn{2}{c|}{(0.048~,~0.048)}&
    \multicolumn{2}{c|}{(0.043~,~0.052)}\cr
    \hline
    \end{tabular}
\end{table*}
\begin{figure*}[ht]
\centering
\includegraphics[width=1.02\textwidth]{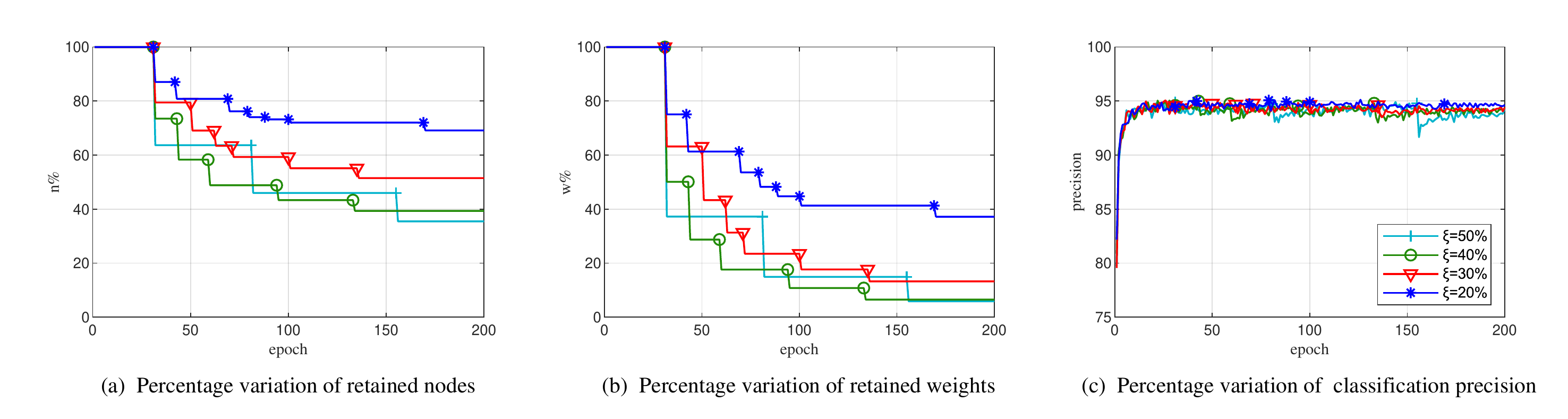}
\caption{Effect of different $\xi$ on our algorithm for pruning CNNs on SVHN.} %图片标题
\end{figure*}

Based on the above empirical results, the detailed summary on our algorithm's performance can be given as follows.
1) To the best of our knowledge, only our algorithm can prune the input layer, since we first
use inverse input autocorrelation matrices $\textbf{P}_t^l$ to evaluate the importance of input channels or nodes.
It is easy to implement and has low computational complexity, since the RLS optimization provides $\textbf{P}_t^l$ naturally.
2) Unlike traditional one-shot pruning algorithms, our algorithm can perform pruning multiple times with negligible loss
 of precision, and can obtain a more reasonable pruning result.
3) Due to the fast convergence of the RLS optimization, our algorithm has high recovery efficency after each pruning.
4) Our algorithm can prune CNNs and FNNs according to the task's difficulty level adaptively, since $\textbf{P}_t^l$  stores rich
input information, which can reflect the difficulty level. For example, a complex image generally requires more important features
to represent it.

\section{Conclusion}
Most of existing pruning algorithms only perform one-shot pruning and don't consider the input information of each layer, so they are generally difficult to obtain a reasonable pruning result. To address this problem, based on RLS optimization, we propose a novel structured pruning algorithm on CNNs, which uses inverse input autocorrelation matrices and weight matrices to evaluate and prune unimportant input channels or nodes layer by layer. Our algorithm
can perform pruning multiple times in a training process. Each pruning only prunes a small portion of channels or nodes to prevent the loss from becoming too large. A new pruning don't be perfomed unitl the performance of the pruned network is recovered.
Therefore, our algorithm can adaptively prune different scale networks to a more reasonable level with negligible loss.
In additon, our algorithm has faster recovery speed and higher pruning efficiency, which are benefiting from the RLS optimization.
Three pruning experiments are performed on MNIST with FNNs, CIFAR-10 with CNNs and SVHN with CNNs.
From the experimental results, it is shown that our algorithm has better pruning performance than other four popular pruning algorithms.
In particular, it is demonstrated that our algorithm can prune CNNs to a reasonable level for fitting the task's difficulty level.

\section*{Declaration of Competing Interest}
The authors declare that they have no known competing financial interests or personal relationships that could have appeared to influence the work reported in this paper.

\section*{Acknowledgments}
This work is supported by the National Natural Science Foundation of China (grant nos. 61762032 and 11961018).



%% References


\bibliographystyle{elsarticle-num}
\bibliography{rlspruning}






% \end{thebibliography}

\end{document}